\documentclass[runningheads]{llncs}

\usepackage{eccv}

\usepackage{eccvabbrv}
\usepackage{multirow} 

\usepackage{graphicx}
\usepackage{booktabs}
\usepackage[utf8]{inputenc}
\usepackage{pifont}
\usepackage{newunicodechar}
\usepackage[table]{xcolor}
\newunicodechar{✓}{\ding{51}}
\newunicodechar{✗}{\ding{55}}
\newcommand{\cmark}{\ding{51}}
\newcommand{\xmark}{\ding{55}}
\newcommand{\ourcheck}{\cellcolor{blue!10}\textcolor{blue!70!black}{\bfseries\cmark}}

\usepackage{latexsym}
\usepackage{amssymb}
\makeatletter
\renewcommand*{\@fnsymbol}[1]{\ensuremath{\ifcase#1\or \star \or \dagger \or \ddagger \or \S \or \P \else \@ctrerr \fi}}
\makeatother

\usepackage[accsupp]{axessibility}  % Improves PDF readability for those with disabilities.

\usepackage{hyperref}

\usepackage{orcidlink}

\begin{document}

\title{Single-Query Person-Centric Bimanual Hand-Object Interaction Detection} 


\author{Jonghyun Kim\thanks{Equal contribution} \and
Junho Roh\textsuperscript{$\star$} \and
Yubin Yoon \and
Hyotae Lee \and
Jongkuk Park \and
Taehwan Hwang \and
Jaechul Kim\thanks{Corresponding author} \and
Jungho Lee\textsuperscript{$\dagger$}
}
\authorrunning{F.~Author et al.}

\institute{AI Lab, CTO Division, LG Electronics, South Korea \\
\email{\{jonghyun0.kim, wesley.roh, yubin.yoon, hyotae.lee, jongkuk.park, mason.hwang, jaechul1220.kim, jungo.lee\}@lge.com}}

\maketitle

\begin{abstract}
Understanding person-level bi-manual interactions requires not only detecting hands, but also identifying which two hands belong to the same person and what each hand interacts with. Existing hand--object interaction methods are mostly \emph{hand-centric}: they treat each hand as an independent instance, which can lead to ambiguous ownership in multi-person scenes.

We propose a \emph{person-centric} formulation in which a \textbf{single query} predicts a structured output for one person, including the human box, body pose, hand boxes and states, and interaction targets. We introduce \textbf{part-aware deformable attention} to allocate attention across human, hand, and pose-specific reference regions, enabling one query to capture the full person structure. We further unify detection and interaction reasoning with a \textbf{hand-to-query relationship matrix}, where each hand selects its interaction target from the detected query set plus a learnable \texttt{off} token, directly recovering the target's box and class without separate object regression.

We build a COCO-based dataset with person-centric bi-manual interaction annotations and define structured metrics for evaluating hand states and complete hand--object tuples. Experiments with a transformer-based detector show that our formulation improves person-level bi-manual interaction parsing and provides an effective unified framework for joint detection, pose estimation, and hand reasoning.
\end{abstract}

\noindent\textbf{Project page:} \url{https://lgecto-ail-vil.github.io/SingleQuery-BHOI/}

\section{Introduction}
\label{sec:intro}

Hands are the primary interface between humans and the physical world. A vision system that can parse \emph{which person} uses \emph{which hand} to manipulate \emph{which object} is essential for fine-grained action understanding, demonstration mining, assistive perception, and robot learning from observation. Recent progress has greatly improved hand detection and hand--object interaction understanding at scale, providing strong supervision and effective baselines for contact reasoning.

A representative line of work formulates hand--object interaction at the \emph{hand level}. 100DOH~\cite{shan2020understanding} predicts, for each hand, its bounding box, hand side, contact state, and in-contact object box. Hands23~\cite{cheng2023towards} further expands the output space with richer interaction signals, including object segments, second objects touched through tools, and grasp/contact categories. These hand-centric formulations are effective for reasoning about individual hands and have substantially advanced large-scale manipulation understanding.

However, they remain limited for \emph{person-level bi-manual interaction parsing}. Existing methods treat hands as independent instances and do not explicitly couple the left and right hands under a shared human instance. In multi-person scenes, this can lead to ambiguous ownership: a model may detect multiple hands and interactions, yet still fail to determine which two hands belong to the same person. More fundamentally, this reflects a mismatch between the \emph{prediction unit} (independent hands) and the \emph{desired interpretation unit} (a person with two hands interacting with objects).

We address this limitation with a \textbf{person-centric structured prediction} framework for bi-manual interaction understanding. Our core idea is to make the \emph{human instance} the primary prediction unit and attach left/right hands as two explicit slots. Concretely, we design a DETR-style set prediction model~\cite{carion2020detr} in which \textbf{a single query} predicts all outputs for one instance, including the human box, body pose, left/right hand boxes, left/right hand states, and hand-specific interaction targets. This formulation parallels person-centric pose estimation, but extends it to structured bi-manual interaction reasoning.

A key challenge is to localize both the human body and small articulated hands within a unified query. To this end, we introduce \textbf{part-aware deformable attention} built upon deformable transformers~\cite{zhu2021deformable}. Each person query uses multiple reference points derived from the predicted human, hand, and pose regions, and decoder heads are distributed across these references. This enables a single query to jointly capture coarse person-level context and fine hand-level details without introducing separate task-specific queries.

\begin{table}[t]
\centering
\caption{Comparison of annotation coverage across hand-contact datasets. 
Our dataset provides richer person-centric supervision, including human boxes, body keypoints, object boxes, and object classes.}
\label{tab:annotation_comparison}
\vspace{-0.3cm}
\resizebox{\textwidth}{!}{
\begin{tabular}{lccccccc}
\toprule
Dataset 
& Person-centric
& Hand box 
& Contact info 
& Object box 
& Object class 
& Human box 
& Body keypoint \\
\midrule
\textbf{Ours}      
& \ourcheck & \cmark & \cmark & \cmark & \ourcheck & \ourcheck & \ourcheck \\
ContactHands \cite{narasimhaswamy2020contacthands}
& \xmark & \cmark & \cmark & \xmark & \xmark & \xmark & \xmark \\
100DOH \cite{shan2020understanding}
& \xmark & \cmark & \cmark & \cmark & \xmark & \xmark & \xmark \\
Hands23 \cite{cheng2023towards}
& \xmark & \cmark & \cmark & \cmark & \xmark & \xmark & \xmark \\
\bottomrule
\end{tabular}
}
\vspace{-0.5cm}
\end{table}

Another challenge is to recover the semantics of the interacted object. Directly regressing an interacting-object box from a person query does not reveal the object's class. We therefore unify \textbf{object detection and interaction inference} in a single model. Each query may represent either a human or an object, and for each hand of a predicted human query, we infer a \textbf{relationship matrix} over the detected query set plus a learnable \texttt{off} token. By selecting an interaction target from the detected queries, the model directly retrieves the target's box and class from the selected query, eliminating the need for separate interacting-object regression.

\begin{figure}[t]
    \centering
    \includegraphics[width=\linewidth]{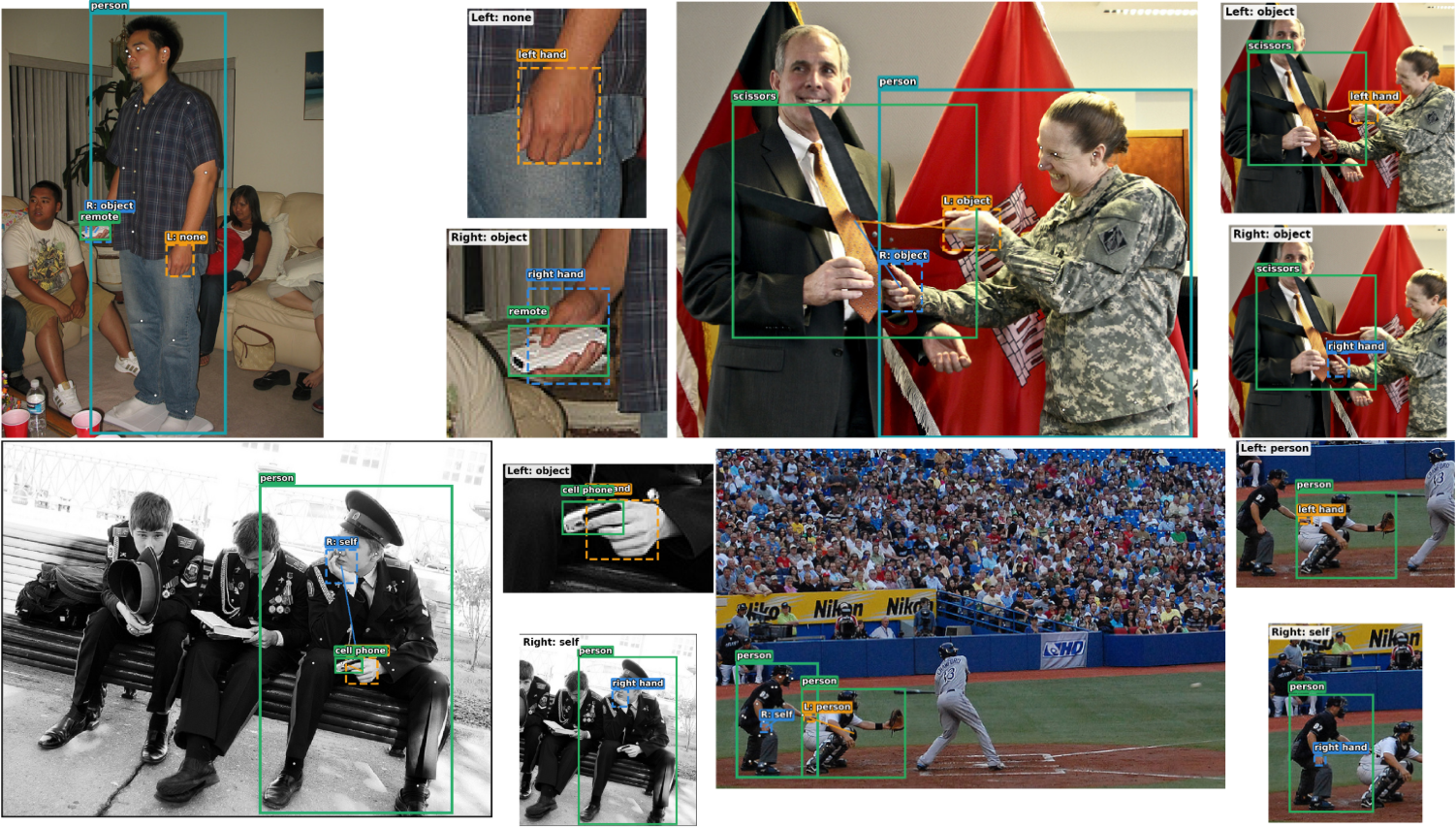}
    \vspace{-0.5cm}
    \caption{\textbf{Visualization examples from our person-centric bi-manual interaction dataset.}
    For each person instance, we annotate a human bounding box, body keypoints, left/right hand regions, hand contact states, and the corresponding interaction targets.
    In each example, the full image is shown together with zoomed-in views of the left and right hands to highlight fine-grained hand-object interactions.
    Dashed boxes indicate hand regions, solid boxes indicate interaction targets, and the annotations explicitly preserve left/right hand ownership under a shared human instance.}
    \vspace{-0.5cm}
    \label{fig:dataset_example}
\end{figure}

To support this task, we construct a COCO-based dataset~\cite{lin2014microsoft} with \textbf{person-centric bi-manual interaction annotations}. Unlike prior hand-centric datasets, our dataset jointly provides hand boxes, hand contact labels, object boxes and classes, as well as human boxes and body keypoints. As shown in Table~\ref{tab:annotation_comparison} and Figure~\ref{fig:dataset_example}, this richer annotation schema is well suited for unified learning of detection, pose estimation, and hand--object interaction reasoning. We further introduce a structured evaluation metric that measures correctness at the level of complete human--hand--object tuples, explicitly penalizing cross-person hand assignment and ownership inconsistencies that are not captured by hand-centric metrics.

We validate the proposed formulation with a transformer-based detector showing that person-centric bi-manual interaction parsing can be both accurate and scalable.

Our contributions are summarized as follows:
\begin{itemize}
    \item We identify a limitation of hand-centric interaction formulations for person-level bi-manual reasoning: they do not explicitly model left/right hand ownership under a shared human instance as a first-class prediction target.
    \item We propose a person-centric single-query framework in which one query predicts all outputs for one instance, jointly solving detection, hand reasoning, and body pose estimation within a unified model.
    \item We introduce part-aware deformable attention and a hand-to-query relationship matrix with a learnable \texttt{off} token, enabling a single query to localize the person and both hands while recovering interaction targets, boxes, and classes from the detected query set.
    \item We build a COCO-based person-centric bi-manual interaction dataset, extending it with a manually labeled validation split, and we propose a structured metric for evaluating complete human--hand--object tuples.
\end{itemize}

% =========================
% Related Work (LaTeX draft)
% =========================
\section{Related Work}
\label{sec:related_work}
\textbf{Hand--object interaction understanding.}
Large-scale hand--object interaction research~\cite{shan2020understanding, cheng2023towards, narasimhaswamy2020contacthands, darkhalil2022visor, bambach2015egohands} has predominantly adopted hand-centric formulations.
100DOH\cite{shan2020understanding} introduced a per-hand output (hand box, side, contact state, and in-contact object box) together with a large dataset for hand-contact understanding.
Hands23\cite{cheng2023towards} further expanded the output space and scale, predicting richer hand interaction cues such as object segments, second objects touched via tools, and contact/grasp types.
While these works significantly advance hand-centric interaction understanding, they do not make \emph{person-level bi-manual ownership consistency} a first-class prediction and evaluation target.
In contrast, our work predicts left/right hands as two structured slots attached to each human instance, explicitly encoding ownership and enabling consistent bi-manual reasoning.
\\
\textbf{Human--object interaction detection.}
The broader HOI literature focuses on detecting human--object pairs, often with action predicates, and has achieved strong performance on standard benchmarks.
Transformer-based methods such as HOTR\cite{kim2021hotr} and HOI Transformer~\cite{zou2021end} unify detection and interaction reasoning in an end-to-end set prediction framework, while subsequent works further improve interaction recognition through stronger predicate context modeling~\cite{zhang2023pvic}.
More recently, open-vocabulary and open-world extensions leverage visual-semantic alignment or multi-modal prompts to generalize beyond closed interaction vocabularies~\cite{lei2025open,yang2024openworld}.
These methods are closely related to our joint detection-and-relation design, but they model interactions at the level of human--object pairs and predicates, rather than \emph{hand-specific} bi-manual interactions within a person-centric structure.
\\
\textbf{Relation prediction over entity sets.}
Predicting relations over detected entities has been widely studied in scene graph generation.
RelTR~\cite{Cong_2022_RelTR} and EGTR~\cite{im2024egtr} model relation reasoning directly over transformer object queries, while recent open-vocabulary methods further introduce vision--language interaction to improve relation generalization beyond closed predicate sets~\cite{min2025vlirm}.
We share the view that relations can be inferred over an entity set; however, our target is not a general predicate graph.
Instead, we predict a compact \emph{hand-to-target selection} for each person query, implemented as a relationship matrix over detected queries, which is specifically tailored to person-centric bi-manual interaction parsing.
\\
\textbf{DETR-style detectors and efficient variants.}
DETR~\cite{carion2020detr} reformulated detection as direct set prediction with bipartite matching, enabling structured outputs without heuristic post-processing.
Deformable DETR~\cite{zhu2021deformable} improves convergence and small-object performance via sparse deformable attention, and DINO~\cite{zhang2023dino} strengthens training with improved denoising and query design.
For practical deployment, RT-DETR~\cite{zhao2024rtdetr} demonstrates real-time end-to-end transformer detection.
Our contribution is orthogonal to these detector advances: rather than proposing a new detector family, we introduce a person-centric representation and a part-aware reference-point allocation that allows a \emph{single query} to jointly reason about the human and both hands, and a relation-based interaction head that leverages the detected entity set to recover interacting objects with classes.
\\
\textbf{Evaluation protocols.}
Existing hand-interaction evaluations primarily measure hand-centric detection and association signals~\cite{shan2020understanding,cheng2023towards}.
Such metrics do not directly penalize ownership inconsistencies (e.g., cross-person hand assignment or left/right slot confusion) when interpreting interactions at the person level.
We therefore propose a structured metric that evaluates complete human--hand--object tuples, aligning evaluation with the person-centric prediction target.

\begin{figure*}[t]
    \centering
    % If needed, you can trim a few pixels of margins:
    % \includegraphics[width=\textwidth,trim=5 5 5 5,clip]{figures/oih_architecture.jpg}
    \includegraphics[width=\textwidth]{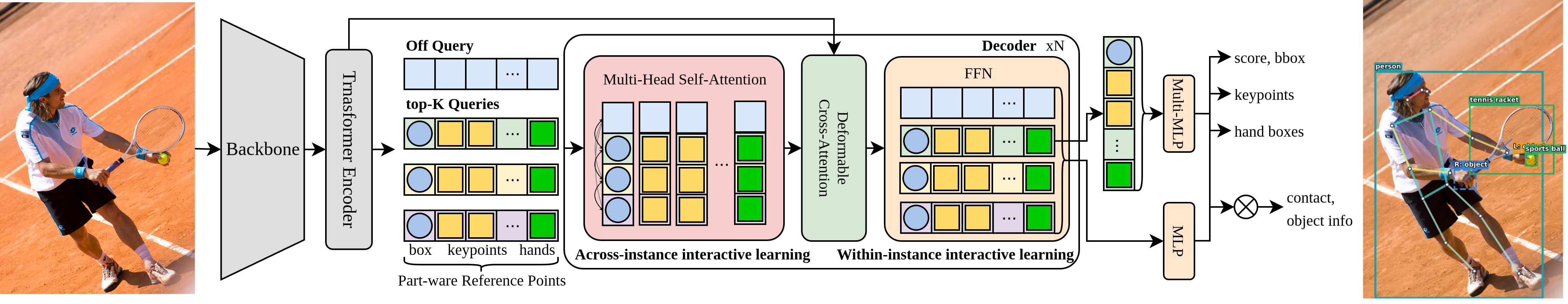}
    \vspace{-0.5cm}
    \caption{\textbf{Model overview.} A backbone and transformer encoder produce top-$K$ queries (plus an \texttt{off} query), which are refined by a deformable decoder with part-aware reference points and interactive learning. The model jointly predicts detection boxes, body keypoints, hand boxes, and hand--target relations to recover contact and interacting object information.}
    \label{fig:architecture}
    \vspace{-0.5cm}
\end{figure*}

% =========================
% Method (LaTeX draft)
% =========================
\section{Method}
\label{sec:method}

\subsection{Task and Person-Centric Output Representation}
\label{sec:task}

Given an input image $\mathbf{x}$, our goal is to predict a set of person-centric interaction tuples in a DETR-style set prediction framework \cite{zhao2024rtdetr, zong2023detrs, zhang2023dino, zhu2021deformable, carion2020detr}.
We use $N$ object queries, where each query can represent either a \textit{human} instance or a general \textit{object} instance.
For every query $i \in \{1,\dots,N\}$, the detector predicts:
(i) a class distribution $\hat{\mathbf{c}}_i \in \mathbb{R}^{C}$ over $C$ categories (including \textit{person}),
and (ii) a bounding box $\hat{\mathbf{b}}_i \in [0,1]^4$.

\paragraph{Person queries and additional heads.}
Each query is processed by the class head, box head, hand-related heads, and pose head.
For queries classified as \textit{person}, these predictions form a structured human representation, including
(i) body keypoints $\hat{\mathbf{K}}_{i} \in [0,1]^{J \times 2}$ for $J$ joints,
(ii) left/right hand boxes $\hat{\mathbf{b}}^{L}_i, \hat{\mathbf{b}}^{R}_i \in [0,1]^4$,
and (iii) left/right hand embeddings $\mathbf{h}^{L}_i, \mathbf{h}^{R}_i \in \mathbb{R}^{d}$ used for interaction reasoning.
Depending on the interaction formulation, a person query further predicts either direct interacting-object boxes or interaction targets through a relation matrix (Sec.~\ref{sec:relation}).

For non-person queries, the same heads are still evaluated during the forward pass.
This is necessary because the predicted hand boxes and pose estimates are also used to construct part-aware reference points in the decoder.
However, hand- and pose-related supervision is applied only to queries matched to human instances.
In other words, non-person queries produce these auxiliary outputs, but no loss is imposed on them.

\paragraph{Hand contact status under two interaction formulations.}
We consider two ways to represent hand--target interactions.

\textbf{(i) Direct regression (DirectBox).}
When directly regressing interacting-object boxes from a person query, the model outputs an additional box $\hat{\mathbf{b}}^{s,\mathrm{obj}}_i$ for each hand side $s\in\{L,R\}$, together with an explicit hand contact status label.
This is necessary because a regressed box alone does not specify whether the hand is in contact, nor the semantic type of the contact.

\textbf{(ii) Relation-based target selection (Relation).}
In our relation-based formulation, each hand selects an interaction target from the detected query set plus a learnable \texttt{off} token.
The hand contact status is then derived from the selected target:
selecting \texttt{off} indicates \textbf{no-contact};
selecting the \emph{self} query ($\hat{t}^s_i = i$) indicates \textbf{self-contact};
and selecting any other query ($\hat{t}^s_i \in \{1,\dots,N\}\setminus\{i\}$) indicates \textbf{contact with another entity} (typically object contact, and it can also represent other-person contact when the selected query corresponds to another person).
This formulation avoids explicit interacting-object box regression and enables retrieving the target box and class directly from the selected query.

\subsection{Part-Aware Deformable Attention with Multiple Reference Points}
\label{sec:partaware}

A key challenge is to localize small articulated hands and body joints while preserving an end-to-end set prediction design.
We address this with a \textbf{part-aware} deformable attention scheme that combines multi-reference sampling with a lightweight \textbf{interactive learning} mechanism.
Prior interactive-learning formulations for structured human prediction often rely on separate query types for instances and parts (e.g., human and keypoint queries)~\cite{yang2023explicit,liu2023group}.
In contrast, our design enables interactive learning \emph{within a single query representation}, without introducing additional task-specific queries.

\paragraph{RT-DETR-style query initialization and refinement.}
We follow RT-DETR~\cite{zhao2024rtdetr} for query preparation and iterative refinement.
After the encoder, we select the top-$K$ candidate queries and apply a box predictor head to obtain initial box hypotheses, which are then used together with the selected query embeddings as inputs to the decoder.
Our part-aware references are built from these initial predictions and are updated across decoder layers using the same refinement mechanism as RT-DETR-style deformable decoding.

\paragraph{Interactive learning within a single query.}
To make one query simultaneously encode global instance context and local part cues, we adopt a head-wise decomposition of the query feature.
Conceptually, different sub-representations of the same query are assigned different structural roles, including the main human body, left/right hands, and body joints.
A lightweight self-attention over queries then enables \emph{interactive learning} at two levels: across queries (instance-to-instance interaction) and within each query (global-to-local coupling across part roles).
Unlike prior methods that realize such interactions through separate instance and part queries, our formulation keeps all roles inside the same query embedding and uses head-wise structure to support interaction.
We defer the full architectural details to the supplementary material.

\paragraph{Part-aware references for all queries.}
At each decoder layer, every query is processed by the box, hand, and pose heads, and the resulting intermediate predictions are used to construct part-aware references for \emph{all} queries.
This is necessary because the decoder relies on these references during iterative refinement.
However, hand- and pose-related supervision is applied only to queries matched to human instances; for non-human queries, the corresponding outputs are produced but not directly supervised.

For query $i$, we derive reference points for the main box and both hand boxes as
\begin{equation}
\mathbf{p}^{H}_i = \phi(\hat{\mathbf{b}}_i), \quad
\mathbf{p}^{L}_i = \phi(\hat{\mathbf{b}}^{L}_i), \quad
\mathbf{p}^{R}_i = \phi(\hat{\mathbf{b}}^{R}_i),
\end{equation}
where $\phi(\cdot)$ maps a box to a 2D reference point (e.g., the box center) in normalized coordinates.

For pose estimation, we further introduce $J$ joint-specific reference regions.
Let $\hat{\mathbf{b}}_i = (\hat{c}^{x}_i, \hat{c}^{y}_i, \hat{w}_i, \hat{h}_i)$ and let $\hat{\mathbf{k}}^{j}_i$ denote the predicted location of joint $j$.
We use $\hat{\mathbf{k}}^{j}_i$ as the joint center and define its width and height by scaling the main box size with learnable parameters:
\begin{equation}
\mathbf{s}^{j}_i =
\left(
\gamma^{j}_{w}\hat{w}_i,\;
\gamma^{j}_{h}\hat{h}_i
\right),
\qquad
\mathbf{p}^{j}_i = (\hat{\mathbf{k}}^{j}_i,\mathbf{s}^{j}_i).
\end{equation}

Decoder attention heads are distributed across these part references, allowing each query to attend jointly to box-level, hand-level, and joint-level regions during refinement.
In practice, this part-aware design is most meaningful for queries that converge to human instances, where interactive learning and part-aware sampling strengthen person-centric reasoning over body structure and hand interactions.

\subsection{Joint Detection and Hand--Object Interaction via Relation Matrix}
\label{sec:relation}

Directly regressing interacting-object boxes makes it difficult to recover the target semantics and to robustly associate targets in crowded scenes.
We therefore couple detection and interaction reasoning by predicting interactions as relations between queries.

\paragraph{Relation embeddings and \texttt{off} token.}
Let $\{\mathbf{z}^{T}_i\}_{i=1}^{N}$ be the final query embeddings.
We project each query embedding to a relation embedding $\mathbf{r}_i \in \mathbb{R}^{d}$ and introduce a learnable \texttt{off} embedding $\mathbf{r}_0 \in \mathbb{R}^{d}$ to represent no-contact.

\paragraph{Hand-to-target scoring.}
For each \textit{person} query $i$, we compute a dot-product score between each hand embedding $\mathbf{h}^{s}_i$ and all target embeddings $\{\mathbf{r}_0,\mathbf{r}_1,\dots,\mathbf{r}_N\}$, followed by a softmax over the $(N+1)$ candidates.
This yields a distribution over interaction targets for each hand side $s\in\{L,R\}$.
The predicted target index $\hat{t}^{s}_i$ is the argmax of this distribution.

\paragraph{Recovering targets and contact types.}
If $\hat{t}^{s}_i = 0$, the hand is predicted as \textbf{no-contact}.
If $\hat{t}^{s}_i = i$, the hand is predicted as \textbf{self-contact}.
Otherwise, the hand is predicted to be in \textbf{contact with another entity} (object or other-person).
Because each query predicts a box and a class distribution, selecting $\hat{t}^{s}_i=j$ directly provides the target's bounding box $\hat{\mathbf{b}}_j$ and semantic class $\arg\max \hat{\mathbf{c}}_j$, without explicit interacting-object regression.

\subsection{Training and Inference}
\label{sec:train_infer}

We instantiate our framework on top of the RT-DETR-style detection pipeline~\cite{zhao2024rtdetr}, which uses top-$K$ query selection after the encoder and iterative decoder refinement from box-aware query initialization.
We train the model end-to-end with DETR-style bipartite matching over human and object instances.
The training objective combines standard detection losses~\cite{rezatofighi2019generalized} with additional supervision for hand boxes, body pose~\cite{maji2022yolo}, and hand--object relations.
In particular, each hand is trained to select its interaction target from the matched query set plus an \texttt{off} option.
The full loss definition and implementation details are deferred to the supplementary material.

At inference time, we keep the top-$K$ queries by detection confidence.
Each query predicts its class and box, and person queries further output body pose, left/right hand boxes, and left/right interaction targets.
If a hand selects a valid target query, the corresponding target box and class are directly retrieved from that query, yielding a person-centric structured prediction.

% =========================
% Dataset (LaTeX draft, updated)
% - k=1.4, l=1.15
% - center-in-expanded-person-box criterion
% - GPT-5 focuses on hand status; ambiguous excluded, clearly wrong refined
% - dataset sizes left as placeholders to be filled later
% - object detection integration with IoU matching + extra "object" class
% =========================
\section{Dataset}
\label{sec:dataset}

\subsection{Overview}
\label{sec:dataset_overview}

We build our dataset on top of COCO~\cite{lin2014microsoft} for two reasons.
First, COCO contains diverse multi-person scenes that are well suited for person-centric bi-manual reasoning.
Second, COCO already provides object detection annotations and human keypoints, allowing us to construct person-level interaction structures without re-annotating all objects from scratch.

To obtain hand annotations, we leverage Hands23~\cite{cheng2023towards}, which provides hand instance annotations for a subset of COCO images.
Our construction pipeline consists of two stages:
(1) converting Hands23 hand instances into \emph{person-centric} left/right hand slots for COCO human instances, and
(2) converting hand contact annotations into interaction targets over COCO entities (objects and persons).
The final dataset supports unified learning of detection, body pose estimation, and person-centric hand interactions.

\subsection{Person--Hand Pair Construction}
\label{sec:dataset_person_hand}

COCO provides human instances with keypoints, including left/right wrist keypoints, while Hands23 provides left/right hand bounding boxes. We align Hands23 hands to COCO persons using wrist-guided geometric matching.
\vspace{-0.2cm}
\paragraph{Candidate generation using visible wrists.}
For each COCO person instance $p$ and hand side $s\in\{L,R\}$, we use the wrist keypoint $\mathbf{w}_{p}^{s}$ when its visibility is non-zero.
For each visible wrist, we consider Hands23 hand boxes $h$ of the same side and compute the distance between the wrist and the hand-box center $\mathbf{c}_{h}$:
\begin{equation}
d(p,h,s) = \|\mathbf{w}^{s}_{p} - \mathbf{c}_{h}\|_2.
\end{equation}
\vspace{-0.5cm}
\paragraph{Geometric matching rule.}
A hand box $h$ is matched to person $p$ for side $s$ if it satisfies both:
(i) the wrist-to-hand-center distance is small relative to the hand size, and
(ii) the hand-box center lies within an expanded region of the person box.
Let $\mathrm{diag}(h)$ denote the diagonal length of $h$, and let $\mathrm{Expand}(\mathbf{b}_{p}; l)$ be the person box expanded by a factor $l$.
We accept a match if
\begin{align}
d(p,h,s) &< k \cdot \mathrm{diag}(h), \quad \text{with } k=1.4, \\
\mathbf{c}_{h} &\in \mathrm{Expand}(\mathbf{b}_{p}; l), \quad \text{with } l=1.15.
\end{align}
\vspace{-0.5cm}
\paragraph{Resolving multiple matches and injecting hand slots.}
If multiple persons satisfy the constraints for the same hand box, we assign the hand to the person with the smallest $d(p,h,s)$.
For each successful match, we inject the Hands23 hand box into the corresponding COCO person instance, producing person-centric records of
\textit{(human box, body keypoints, left/right hand boxes)}.
\vspace{-0.3cm}
\paragraph{LLM-assisted verification.}
Because the above alignment is rule-based, incorrect mappings can still occur under severe occlusion or when multiple people overlap.
To reduce such noise, we apply an additional verification step after the Hands23-to-COCO mapping.
The verifier uses (i) the full image, (ii) a local crop around the hand and its surrounding context, and (iii) a relative depth map extracted by DepthAnythingV2~\cite{yang2024depth}.
These complementary cues provide spatial evidence for occlusion and front--back ordering, helping to identify ambiguous or clearly incorrect matches.
Such cases are filtered or corrected.
Detailed verification procedures are provided in the supplementary material.
\vspace{-0.1cm}

\subsection{Constructing Interaction Targets}
\label{sec:dataset_targets}

Our model represents interactions via target selection over the entity set plus an \texttt{off} option.
Accordingly, we convert hand-contact annotations into interaction targets over COCO entities.

\paragraph{Raw contact taxonomy.}
We retain four contact categories:
\texttt{no\_contact}, \texttt{self\_contact}, \texttt{other\_person\_contact}, and \texttt{object\_contact}.
For the direct-regression baseline and hand-state evaluation, these labels can be collapsed into a binary state (\texttt{none} vs.\ \texttt{hold}), where \texttt{no\_contact}$\rightarrow$\texttt{none} and all others$\rightarrow$\texttt{hold}.
For the relation-based model, we preserve the target semantics through explicit target assignment.

\paragraph{IoU-based target assignment.}
Hands23 provides a contact-associated interaction box for each contacting hand.
We match this interaction box to COCO entities in the same image using IoU ($\ge 0.5$).
For \texttt{object\_contact}, we compare it against COCO object boxes and assign the highest-IoU object.
For \texttt{other\_person\_contact}, we compare it against COCO person boxes other than the current person and assign the highest-IoU person.
For \texttt{self\_contact}, the target is the current person instance, and for \texttt{no\_contact}, the target is \texttt{off}.
This produces interaction targets compatible with our relation-based head, where each hand selects one target from the detected entity set plus \texttt{off}.

\paragraph{Handling unmatched targets.}
If no COCO object matches the interaction box for \texttt{object\_contact}, we create a pseudo target box using the interaction box and assign it a generic class \texttt{object}, used only as an interaction placeholder during training.
If no valid person match is found for \texttt{other\_person\_contact}, the sample is discarded to avoid ambiguous supervision.

\subsection{Validation Set Construction}
\label{sec:dataset_val}

Hands23 does not overlap with COCO val2017 in our setting.
We therefore manually annotate a subset of COCO val2017 images containing at least one person (approximately 2K images) to create a human-verified validation split.
These annotations follow the same person-centric schema as the training set, including left/right hand boxes and contact labels, and interaction targets are constructed using the same target-assignment rules described above.

\subsection{Annotation reliability.}
Since our training labels are generated by rule-based COCO--Hands23 alignment, annotation noise may arise under occlusion, overlap, or severe scale variation. To improve label reliability, we apply a conservative VLM-based verification stage using the full image, target person crop, and target hand crop, each paired with a DepthAnythingV2\cite{yang2024depth} relative depth map. The verifier follows a strict no-guessing policy and outputs \texttt{contact}, \texttt{no-contact}, or \texttt{uncertain}; uncertain cases are excluded from training.

We further conduct a manual audit on 770 samples. The uncertain ratio is 1.3\%, and the annotation error rate decreases from 2.5\% for the rule-based labels to 2.1\% after verification. Two annotators also double-labeled 500 samples and achieved 98.4\% agreement. The remaining errors are mostly due to visually ambiguous cases such as blurry or low-resolution hands, very small hands, occlusion, and depth-hard configurations.

\section{Evaluation}
\label{sec:eval}

We evaluate person-centric bi-manual interaction parsing with three metrics: \textit{soft}, \textit{medium}, and \textit{hard}.
All three metrics first match predicted and ground-truth person instances using one-to-one IoU-based assignment~\cite{carion2020detr}, and retain only pairs whose person-box IoU exceeds a threshold $\tau_p$ (we use $\tau_p=0.5$, following standard detection overlap criteria~\cite{everingham2010pascal}).

\paragraph{Notation.}
Let $\mathcal{G}$ and $\mathcal{P}$ denote the sets of ground-truth and predicted person instances, and let $\mathcal{A}\subseteq\mathcal{G}\times\mathcal{P}$ be the set of matched person pairs.
For each matched pair $(g,p)\in\mathcal{A}$ and hand side $s\in\{L,R\}$, let $y_g^s\in\{\texttt{hold},\texttt{none}\}$ be the ground-truth hand state, $\hat{y}_p^s$ the predicted hand state, and $m_g^s\in\{0,1\}$ an availability mask indicating whether the hand annotation is present.

\paragraph{Soft metric.}
The soft metric measures hand-state accuracy over matched persons:
\begin{equation}
\mathrm{Acc}_{\mathrm{soft}}
=
\frac{
\sum_{(g,p)\in\mathcal{A}}
\sum_{s\in\{L,R\}}
m_g^s\cdot \mathbb{1}\left[\hat{y}_p^s = y_g^s\right]
}{
\sum_{(g,p)\in\mathcal{A}}
\sum_{s\in\{L,R\}}
m_g^s
}.
\end{equation}

\paragraph{Medium and hard metrics.}
The \textbf{medium metric} additionally requires correct interacted-object localization when the ground-truth state is \texttt{hold}, following HOI-style protocols that require both human and object localization to be correct~\cite{chao2018learning,gupta2015visual}.
The \textbf{hard metric} further requires correct hand localization, i.e., the predicted hand box must also overlap the ground-truth hand box with IoU above a threshold.
By default, we use $\tau_o=0.5$ for object-box IoU and $\tau_h=0.5$ for hand-box IoU.
In the main paper, we report the average over left and right hands, while full formal definitions and left/right breakdowns are provided in the supplementary material.

% =========================
% Experiments
% =========================
\section{Experiments}
\label{sec:experiments}

\subsection{Experimental Setup}
\label{sec:exp_setup}

\paragraph{Datasets.}
We train and evaluate our models on the proposed COCO-based person-centric bi-manual interaction dataset.
The training set is constructed by aligning COCO human instances and keypoints with hand annotations.
The validation set consists of manually annotated COCO val2017 images, where the original COCO annotations are used for detection and pose evaluation, and the manual hand annotations are used for hand-interaction evaluation.

To study the effect of additional hand-centric supervision, we further convert 100DOH~\cite{shan2020understanding} into our unified training format using the same person--hand merging pipeline.
Since 100DOH does not provide object detection or body pose annotations, we generate pseudo labels for these tasks using off-the-shelf object detection \cite{zong2023detrs} and pose estimation \cite{geng2023human} models, and use the converted 100DOH samples as additional training data.

We also evaluate on ContactHands~\cite{narasimhaswamy2020contacthands}, using its full dataset as an evaluation set.
Unlike our primary benchmark, ContactHands provides only hand-level contact annotations, i.e., hand states and hand bounding boxes, without interacted object annotations.
As a result, we report only the soft hand-state accuracy $\mathrm{Acc}_{\mathrm{soft}}$.
To incorporate ContactHands into our evaluation pipeline, we apply the similar data conversion strategy used for 100DOH: missing labels required by our formulation are supplemented with pseudo annotations.
This allows us to assess whether the proposed person-centric formulation generalizes to an additional hand-contact benchmark even when only partial supervision is available.

\paragraph{Implementation details.}
Unless otherwise specified, all main experiments use the RT-DETR-R50-m variant as the base detector, on top of which we add the proposed person-centric hand, pose, and relation heads.

\paragraph{Evaluation metrics.}
We evaluate three tasks: object detection, body pose estimation, and hand--object interaction reasoning. We report mAP and AP for object detection and body pose estimation respectively. For hand reasoning, we use the two person-matched metrics defined in Sec.~\ref{sec:eval}. For both hand metrics, we report left-hand, right-hand, and average scores.

% \begin{table}[t]
% \centering
% \caption{Effect of relation-based interaction modeling. We compare direct interacting-object box regression (\textbf{DirectBox}) and the relation-based method (\textbf{Relation}).}
% \label{tab:relation_only}
% \begin{tabular}{l|cccc|c}
% \toprule
% \multirow{2}{*}{Method}
% & \multicolumn{4}{c|}{COCOval2017}
% & \multicolumn{1}{c}{ContactHands} \\
% \cmidrule(lr){2-5} \cmidrule(lr){6-6}
% & Det. mAP & Soft-Avg & Mid-Avg & Hard-Avg
% & Soft-Avg \\
% \midrule
% DirectBox & 49.1  & 81.5 & 44.9 & 12.6 & 80.7 \\
% Relation  & 48.8  & 80.0 & 60.7 & 28.9 & 81.7 \\
% \bottomrule
% \end{tabular}
% \end{table}

\subsection{Ablation Studies}
\paragraph{Effect of relation-based interaction modeling.}
\label{sec:exp_relation}
Table~\ref{tab:relation_pose_ablation} compares direct interacting-object box regression (\textbf{DirectBox}) and the proposed relation-based target selection (\textbf{Relation}) under the same training setting without pose supervision.
Although both variants achieve similar detection performance on COCOval2017 (49.1 vs.\ 48.8 mAP), relation-based modeling yields substantially stronger interaction reasoning: $\mathrm{Acc}_{\mathrm{mid}}$ improves from 44.9 to 60.8 (+15.9) and $\mathrm{Acc}_{\mathrm{hard}}$ improves from 12.6 to 28.9 (+16.3).
This indicates that selecting interaction targets from the detected query set is considerably more reliable than directly regressing interacting-object boxes, especially under the stricter tuple correctness criteria that require accurate localization.
Interestingly, the relation-based model shows a small drop in $\mathrm{Acc}_{\mathrm{soft}}$ on COCO (81.5 to 80.0), suggesting that relation modeling primarily benefits target association and localization rather than hand-state classification alone.
On ContactHands, where only hand-state supervision is evaluable, the relation model slightly improves $\mathrm{Acc}_{\mathrm{soft}}$ (80.7 to 81.7), indicating that relation-based training does not harm state recognition and may improve generalization.

\begin{table*}[t]
\centering
\caption{Comparison of direct regression and relation-based formulations, with and without joint body pose learning. This table summarizes the effects of both relation modeling and pose supervision on detection, pose estimation, and hand-interaction performance.}
\vspace{-0.3cm}
\label{tab:relation_pose_ablation}
\resizebox{\textwidth}{!}{
\begin{tabular}{lcc|ccccc|c}
\toprule
\multirow{2}{*}{Method}
& \multirow{2}{*}{Rel.}
& \multirow{2}{*}{Pose}
& \multicolumn{5}{c|}{COCOval2017}
& \multicolumn{1}{c}{ContactHands} \\
\cmidrule(lr){4-8} \cmidrule(lr){9-9}
& &
& Det. mAP & Pose AP & $\mathrm{Acc}_{\mathrm{soft}}$ & $\mathrm{Acc}_{\mathrm{mid}}$ & $\mathrm{Acc}_{\mathrm{hard}}$
& $\mathrm{Acc}_{\mathrm{soft}}$ \\
\midrule
DirectBox      & ✗ & ✗ & 49.1 & N/A  & 81.5 & 44.9 & 12.6 & 80.7 \\
DirectBox+Pose & ✗ & ✓ & 47.7 & 61.8 & 83.0 & 57.6 & 30.1 & 81.3 \\ \hline
Relation       & ✓ & ✗ & 48.8 & N/A  & 80.0 & 60.8 & 28.9 & 81.7 \\
Relation+Pose  & ✓ & ✓ & 47.1 & 62.3 & 83.8 & 64.6 & 31.3 & 82.2 \\
\bottomrule
\end{tabular}
}
\vspace{-0.5cm}
\end{table*}

\paragraph{Effect of joint body pose learning.}
\label{sec:exp_pose}
Adding body pose supervision consistently improves hand reasoning for both formulations.
For direct regression, \textbf{DirectBox+Pose} boosts $\mathrm{Acc}_{\mathrm{mid}}$ from 44.9 to 57.6 (+12.7) and $\mathrm{Acc}_{\mathrm{hard}}$ from 12.6 to 30.1 (+17.5), while also improving $\mathrm{Acc}_{\mathrm{soft}}$ from 81.5 to 83.0.
For the relation-based model, \textbf{Relation+Pose} further improves $\mathrm{Acc}_{\mathrm{mid}}$ from 60.8 to 64.6 (+3.8) and $\mathrm{Acc}_{\mathrm{hard}}$ from 28.9 to 31.3 (+2.4), and yields the best overall $\mathrm{Acc}_{\mathrm{soft}}$ on COCO (83.8) and ContactHands (82.2).
These gains support our hypothesis that pose supervision strengthens person-centric structural understanding, which in turn improves hand localization and interaction reasoning.
We note that incorporating pose slightly decreases detection mAP (e.g., 49.1$\rightarrow$47.7 for DirectBox and 48.8$\rightarrow$47.1 for Relation), which is consistent with the task-conflict phenomenon commonly observed in multi-task learning~\cite{deng2023split,liu2021conflict,yu2020gradient}; nevertheless, the consistent improvements in $\mathrm{Acc}_{\mathrm{mid}}$ and $\mathrm{Acc}_{\mathrm{hard}}$ demonstrate that joint pose learning is beneficial for the core person-centric interaction objective.

\begin{table}[t]
\centering
\caption{Effect of adding 100DOH supervision to the Relation+Pose model.}
\vspace{-0.3cm}
\label{tab:doh_relation_pose}
\begin{tabular}{lc|ccccc|c}
\toprule
\multirow{2}{*}{Method}
& \multirow{2}{*}{+100DOH}
& \multicolumn{5}{c|}{COCOval2017}
& \multicolumn{1}{c}{ContactHands} \\
\cmidrule(lr){3-7} \cmidrule(lr){8-8}
& 
& Det. mAP & Pose AP & $\mathrm{Acc}_{\mathrm{soft}}$ & $\mathrm{Acc}_{\mathrm{mid}}$ & $\mathrm{Acc}_{\mathrm{hard}}$
& $\mathrm{Acc}_{\mathrm{soft}}$ \\
\midrule
Relation+Pose & ✗ & 47.1 & 62.3 & 83.8 & 64.6 & 31.3 & 82.2 \\
Relation+Pose & ✓ & 48.6 & 63.2 & 84.0 & 66.2 & 32.7 & 81.2 \\
\bottomrule
\end{tabular}
\vspace{-0.5cm}
\end{table}

\paragraph{Effect of additional 100DOH supervision.}
Table~\ref{tab:doh_relation_pose} evaluates whether the proposed Relation+Pose model can benefit from additional hand-centric supervision from 100DOH.
Adding 100DOH consistently improves performance on COCOval2017, increasing detection mAP from 47.1 to 48.6 (+1.5), pose AP from 62.3 to 63.2 (+0.9), $\mathrm{Acc}_{\mathrm{soft}}$ from 83.8 to 84.0 (+0.2), $\mathrm{Acc}_{\mathrm{mid}}$ from 64.6 to 66.2 (+1.6), and $\mathrm{Acc}_{\mathrm{hard}}$ from 31.3 to 32.7 (+1.4).
Notably, the gains are substantially larger on $\mathrm{Acc}_{\mathrm{mid}}$ and $\mathrm{Acc}_{\mathrm{hard}}$ than on $\mathrm{Acc}_{\mathrm{soft}}$.
This is consistent with the annotation structure of 100DOH, which provides explicit interacted-object bounding boxes and therefore supplies direct supervision for hand--object localization and association, rather than only hand-state recognition.
In other words, 100DOH primarily strengthens the model's ability to recover complete hand--object tuples, which is more directly reflected in the stricter medium and hard metrics.

On ContactHands, by contrast, the model shows a slight drop in $\mathrm{Acc}_{\mathrm{soft}}$ (82.2 to 81.2).
A likely reason is that ContactHands evaluates only hand-state recognition, while the additional 100DOH supervision mainly benefits target-aware interaction reasoning.
This suggests that the added hand-centric supervision is more aligned with our tuple-level COCO evaluation than with hand-state-only transfer to ContactHands.

\paragraph{Effect of part-aware reference allocation.}
Table~\ref{tab:partaware_ablation} evaluates the impact of the proposed part-aware reference allocation.
Without part-aware allocation, all decoder heads share the main box reference, which favors instance-level localization and yields slightly higher detection mAP (e.g., 48.7 vs.\ 47.7 for DirectBox+Pose and 48.6 vs.\ 47.1 for Relation+Pose).
However, this shared-reference scheme severely hurts part-level reasoning: pose AP drops to $\sim$42 for both formulations, and the strict hand metrics degrade dramatically.

In contrast, enabling part-aware allocation introduces joint- and hand-specific references, which substantially improves body pose and hand-interaction performance despite a small reduction in detection mAP.
For \textbf{DirectBox+Pose}, pose AP improves from 43.4 to 61.8 (+18.4), and hand performance increases across all metrics, with particularly large gains on tuple-level correctness: $\mathrm{Acc}_{\mathrm{mid}}$ rises from 43.1 to 57.6 (+14.5) and $\mathrm{Acc}_{\mathrm{hard}}$ from 11.5 to 30.1 (+18.6).
This suggests that direct interacting-object regression is highly sensitive to whether the decoder attends to the correct local regions around hands and joints.

For \textbf{Relation+Pose}, part-aware allocation yields similar pose gains (42.8$\rightarrow$62.3, +19.5) and improves hand metrics as well.
Because the relation-based formulation retrieves interaction targets from detected queries, $\mathrm{Acc}_{\mathrm{mid}}$ changes only modestly (63.6$\rightarrow$64.6, +1.0), reflecting its relative robustness to moderate detection variation.
Nevertheless, the stricter $\mathrm{Acc}_{\mathrm{hard}}$ improves dramatically (10.3$\rightarrow$31.3, +21.0), indicating that accurate hand localization and part-level attention are crucial once the evaluation requires precise hand boxes in addition to correct interaction targets.
On ContactHands, part-aware allocation provides consistent gains for both formulations (80.5$\rightarrow$81.3 and 81.7$\rightarrow$82.2).

Overall, these results confirm the intended trade-off: allocating all heads to the main box reference can slightly favor detection, whereas part-aware reference allocation is essential for learning accurate body pose and reliable person-centric bi-manual interactions, under stricter hand-centric localization criteria.

\begin{table*}[t]
\centering
\caption{Effect of part-aware reference allocation. For pose-enabled models, part-aware attention additionally includes joint-specific reference regions.}
\vspace{-0.3cm}
\label{tab:partaware_ablation}
\resizebox{\textwidth}{!}{
\begin{tabular}{lc|ccccc|c}
\toprule
\multirow{2}{*}{Method}
& \multirow{2}{*}{PartAware}
& \multicolumn{5}{c|}{COCOval2017}
& \multicolumn{1}{c}{ContactHands} \\
\cmidrule(lr){3-7} \cmidrule(lr){8-8}
&
& Det. mAP & Pose AP & $\mathrm{Acc}_{\mathrm{soft}}$ & $\mathrm{Acc}_{\mathrm{mid}}$ & $\mathrm{Acc}_{\mathrm{hard}}$
& $\mathrm{Acc}_{\mathrm{soft}}$ \\
\midrule
DirectBox+Pose & ✗ & 48.7 & 43.4 & 80.4 & 43.1 & 11.5  & 80.5 \\
DirectBox+Pose & ✓ & 47.7 & 61.8 & 83.0 & 57.6 & 30.1 & 81.3 \\ \hline
Relation+Pose  & ✗ & 48.6 & 42.8 & 83.0 & 63.6 & 10.3  & 81.7 \\
Relation+Pose  & ✓ & 47.1 & 62.3 & 83.8 & 64.6 & 31.3 & 82.2 \\
\bottomrule
\end{tabular}
}
\vspace{-0.5cm}
\end{table*}

\subsection{Crowding Analysis}

We further analyze model robustness under ownership ambiguity by splitting COCO val2017 according to the number of people in each image. As shown in Table~\ref{tab}, performance is relatively stable for single-person and moderately crowded scenes, but drops in highly crowded scenes with $\geq$5 people. This suggests that severe occlusion, hand overlap, and small visible hand regions remain the main challenges in extreme crowding.

\begin{table}[t]
\centering
\caption{\textbf{Crowding split of the final model on COCO val2017.}}
\label{tab}
\small
\setlength{\tabcolsep}{4pt}
\renewcommand{\arraystretch}{0.95}
\begin{tabular}{lccc}
\toprule
\# persons & Acc${\mathrm{soft}}$ & Acc${\mathrm{mid}}$ & Acc$_{\mathrm{hard}}$ \\
\midrule
1 & 90.28 & 79.42 & 54.00 \\
2--4 & 86.58 & 67.31 & 33.09 \\
$\geq$5 & 75.84 & 56.17 & 22.02 \\
\bottomrule
\end{tabular}
\end{table}

\section{Conclusion}
\label{sec:conclusion}

We presented a person-centric framework for bi-manual interaction understanding that explicitly couples the left and right hands under a shared human instance, addressing a key limitation of prior hand-centric formulations in multi-person scenes.
Our approach represents each instance with a single query that jointly supports object detection, body pose estimation, and hand interaction reasoning.
To enable accurate part-level localization without introducing task-specific queries, we proposed part-aware deformable attention with multiple reference regions, and we introduced a relation-based interaction head that selects hand targets from the detected query set with an \texttt{off} option, directly recovering target boxes and classes from the selected queries.

To support this setting, we constructed a COCO-based dataset with person-centric annotations and proposed structured evaluation metrics that progressively measure hand-state accuracy and tuple correctness under increasingly strict localization requirements.
Experiments demonstrate that relation-based target selection substantially improves tuple-level performance over direct interacting-object regression, joint pose learning strengthens person-centric hand reasoning, and part-aware reference allocation is critical for accurate pose and hand localization, especially under the strictest evaluation criterion.
We also showed that additional hand-centric supervision from 100DOH further improves target-aware interaction performance.

We believe this work provides a practical and scalable direction for holistic human-centric understanding.
Future work includes expanding the dataset to broader domains and interaction types, improving robustness under severe occlusion and crowded scenes, and extending the formulation to richer interaction semantics beyond contact-level reasoning.

\clearpage  % TODO FINAL: This \clearpage needs to be removed from both review and camera-ready versions.

\section*{Acknowledgements}
% Please insert your acknowledgments here.

\bibliographystyle{splncs04}
\bibliography{main}

\end{document}